\documentclass[10pt, a4paper]{article}

\usepackage{booktabs}
\usepackage{hyperref}
\usepackage{tabularx}
\usepackage{paralist}
\usepackage{graphicx}
\usepackage{pgffor}
\usepackage{xcolor}
\usepackage{xspace}
\usepackage{multirow}
\usepackage[dvipsnames,table]{xcolor}

\newcommand{\gd}[1]{\medmuskip=0mu\relax\textcolor{OliveGreen}{\scriptsize $\Delta+#1$}}

\newcommand{\rd}[1]{\medmuskip=0mu\relax\textcolor{Maroon}{\scriptsize $\Delta-#1$}}

\newcommand{\contarga}{\textsc{ContArgA}\xspace}
\newcommand{\F}{F$_1$\xspace}
\newcommand{\emohi}[1]{\textit{#1}\xspace}

\newcommand{\anger}{\emohi{anger}}

\newcommand{\joy}{\emohi{joy}}

\newcommand{\fear}{\emohi{fear}}

\newcommand{\sadness}{\emohi{sadness}}

\newcommand{\disgust}{\emohi{disgust}}

\newcommand{\pride}{\emohi{pride}}

\newcommand{\trust}{\emohi{trust}}

\newcommand{\familiarity}{\emohi{familiarity}}

\newcommand{\pleasantness}{\emohi{pleasantness}}

\newcommand{\unpleasantness}{\emohi{unpleasantness}}

\newcommand{\negconseq}{\emohi{negative consequentiality}}

\newcommand{\conseqmanage}{\emohi{consequence manageability}}

\newcommand{\mistral}{\texttt{Mistral}\xspace}
\newcommand{\gemma}{\texttt{Gemma}\xspace}
\newcommand{\llama}{\texttt{Llama}\xspace}

\usepackage[final]{lrec2026} 

\title{Categorical Emotions or Appraisals -- Which Emotion Model Explains
  Argument Convincingness Better?}

\name{Lynn Greschner, Meike Bauer, Sabine Weber, Roman Klinger} 

\address{Fundamentals of Natural Language Processing, University of Bamberg, Germany\\
  \{firstname.lastname\}@uni-bamberg.de\\}

\abstract{%
  The convincingness of an argument does not only depend on its
  structure (logos), the person who makes the argument (ethos), but
  also on the emotion that it causes in the recipient (pathos). While
  the overall intensity and categorical values of emotions in
  arguments have received considerable attention in the research
  community, we argue that the emotion an argument evokes in a
  recipient is subjective. It depends on the recipient's goals,
  standards, prior knowledge, and stance. Appraisal theories lend
  themselves as a link between the subjective cognitive assessment of
  events and emotions. They have been used in event-centric emotion
  analysis, but their suitability for assessing argument convincingness remains unexplored. In
  this paper, we evaluate whether appraisal theories are suitable for
  emotion analysis in arguments by considering subjective cognitive
  evaluations of the importance and impact of an argument on its
  receiver. Based on the annotations in the recently published
  \contarga corpus, we perform zero-shot prompting experiments to
  evaluate the importance of gold-annotated and predicted emotions and
  appraisals for the assessment of the subjective convincingness
  labels. We find that, while categorical emotion information does
  improve convincingness prediction, the improvement is more
  pronounced with appraisals. This work presents the first systematic comparison between emotion models for convincingness prediction, demonstrating the advantage of appraisals, providing insights for theoretical and practical applications in computational argumentation.
  \\%
  \newline %
}

\begin{document}

\maketitleabstract

\section{Introduction}
The analysis of arguments and their quality, persusasiveness and
convincingness received substantial attention
\citep{lawrence-reed-2019-argument}. Argument quality assessment contains various
subtasks, including quantifying the logical, 
rhetorical, and dialectical quality of the argument \citep[cited after
\newcite{wachsmuth-etal-2024-argument}]{Blair2011-BLAGIT}.

An important aspect of dialectical quality is the convincingness of an
argument, which we focus on in this paper. The convincingness of an argument is distinct from the overall effectiveness of argumentation because of its inherently subjective nature.  Part of the subjective
evaluation of an argument regarding its convincingness is the
emotional appeal -- changing the emotional state of a receiver such
that they are more open to the argument.  Most work handled
\textit{emotional appeal} as a continuous score or a binary variable
\citep{wachsmuth-etal-2024-argument}. For instance,
\newcite{chen-eger-2025-emotions} show that manipulating the emotional
appeal in a given argument changes its convincingness. Other studies
focus on discrete emotions in
arguments. \newcite{greschner-klinger-2025-fearful} do however show
that emotion recognition in arguments is a particularly challenging
task, with low performance scores.

We hypothesize that this is because emotions in arguments develop in
context of the argument recipient, including their demographic and psychological
traits and states, their prior world knowledge and experiences, and
stances towards topics. In general emotion analysis tasks, this subjective cognitive evaluation has been approached with the help of
appraisal theories. Appraisal theories describe the cognitive
evaluation of an event and the relationship of this evaluation to concrete emotion
categories. \newcite{smith1985patterns-appraisals}, for instance, show
that six appraisal variables explain 15 discrete emotions, namely (1)
how pleasant an event is (pleasantness, likely to be associated with
joy, but unlikely to appear with disgust), (2) how much effort an
event can be expected to cause (anticipated effort, likely to be high
when anger or fear is experienced), (3) how certain the experiencer is
in a specific situation (certainty, low, e.g., in the context of hope
or surprise), (4) how much attention is devoted to the event
(attention, likely to be low, e.g., in the case of boredom or
disgust), (5) how much responsibility the experiencer of the emotion
holds for what has happened (self-other responsibility/control, high
for feeling challenge or pride), and (6) how much the experiencer has
control over the situation (situational control, low in the case of
anger). \citet{scherer2001appraisal} points out the sequential nature
of the cognitive evaluation of an event: A person first decides its
relevance, its goal conduciveness, one's own ability to cope with
outcomes, and the internal and external norms (regarding moral aspects
and the legal situation).

Appraisal theories find application in natural language processing:
\newcite{hofmann-etal-2020-appraisal} annotate the
appraisal evaluation that somebody experiences throughout an
event. \newcite{Troiano2023-crowdenvent} introduce a novel annotation
framework, in which the potential noise is reduced by
retrieving appraisal variables directly from the author of a text who
lived through an event. Appraisals have also been used to study Reddit
\citep{stranisci-etal-2022-appreddit}, to explain conspiracy theories
\citep{pummerer_appraisal_2024}, to understand the information
processing in large language models \citep{zhan-etal-2023-evaluating},
to tailor emotion assignments to particular entities
\citep{troiano-xenvent}, or to study emotion inference
\citep{tak-etal-2025-mechanistic}.

This diversity shows how versatile the approach is. Nevertheless, appraisals
have not been used to study arguments yet, despite the fact that the
emotion that somebody develops based on an argument does depend
on a subjective cognitive evaluation. Using the \contarga{} corpus \citep{greschner2025trustmeiconvince}, which includes arguments annotated for emotions and appraisals, we explore whether appraisals can computationally explain perceived convincingness. We compare this setup to the use of categorical
emotion models.

Our concrete research questions are:
\begin{itemize}
    \item RQ1: Which emotion model helps LLMs to improve convincingness predictions?
    \item RQ2: Does jointly predicting emotions/appraisals and convincingness improve the performance compared to the single task predictions?
\end{itemize}

Our results demonstrate that both emotion models improve the convincingness prediction across three models, with appraisals providing stronger improvements than categorical emotions. The joint prediction consistently underperforms pipeline approaches.

\section{Related Work}

\subsection{Emotion Analysis and Appraisals}\label{sec:rel_work_appraisals}
Emotion analysis in NLP commonly builds on top of two main types of psychological emotion models: categorical
and dimensional approaches. The most commonly used
categorical model in NLP is Ekman's basic emotion model, which
proposes that
six discrete emotions are universal regarding stimulus events and
reactions, namely anger, surprise, disgust, joy, fear, and
sadness \citep{ekman1992-emotions}. In contrast to such categorical
models, dimensional models represent emotions along
continuous axes in multidimensional spaces. One widely used example in
NLP is the Circumplex Model of Affect \citep{posner2005affect}, where
emotions are evaluated in terms of valence and arousal.

Appraisal theories are approaches that received attention in NLP only
more recently. They provide access to emotions through cognitive
evaluations of events \citep{scherer2001appraisal}. There are multiple
frameworks of appraisal theories
\citep{roseman1984cognitive,roseman2001appraisal,scherer2009dynamic},
which propose varieties of appraisal variables. Therefore, also
specialized frameworks have been proposed, including one for
evaluating arguments \citep{greschner2025trustmeiconvince}. Another
theory has been developed particularly for the analysis of conspiracy
theories \citep{pummerer_appraisal_2024}. While there is no one set of
appraisal variables, common variables encompass aspects of agency,
pleasantness, consequences on the self, responsibility, expected
effort, and novelty.

Categorical emotion models have received more attention in NLP than
appraisal theories, but there is now also considerable work that
employed appraisal theories. Examples include work for the study of
coping strategies \citep{troiano-etal-2024-dealing}, social media
analysis \citep{stranisci-etal-2022-appreddit}, or emotion event self
reports
\citep{hofmann-etal-2020-appraisal}. \citet{Troiano2023-crowdenvent}
propose the largest set of appraisal variables for emotion analysis of
events (Crowd-enVent). Using this corpus,
\citet{tak-etal-2025-mechanistic} probe models to understand whether
they process emotions similarly to
humans. \citet{yeo-jaidka-2025-beyond} train an appraisal predictor on
the corpus and apply it to conversations, capturing changes in
emotional states throughout them. \citet{yeo-jaidka-2025-beyond} use
appraisal theories as the theoretical groundwork for a theory-of-mind
framework-based dataset to assess inferred emotions (from context),
comparing humans and LLMs, highlighting the need of psychological
theories for evaluation LLMs on emotion (reasoning) tasks.

Recently, \citet{greschner2025trustmeiconvince} expand and adapt the
annotation scheme of the Crowd-enVent corpus
\citep{Troiano2023-crowdenvent} to the analysis of arguments. However,
the authors do not conduct a modeling study that would provide
insights into how convincingness modeling performs under specific
argument appraisal or emotional contexts. We fill this gap and make
their novel corpus the data source for our investigation.

The convincingness of arguments is inherently subjective, depending on how receivers cognitively evaluate it. This suggests that appraisal theories, with their focus on subjective cognitive evaluation, may be particularly well-suited for computationally assessing argument convincingness – a hypothesis we explore in this work.

\subsection{Arguments and Convincingness}
Assessing the quality, persuasiveness, and convincingness of an
argument are closely related tasks in the field of argument mining,
which have received considerable attention
\citep{lawrence-reed-2019-argument, habernal-gurevych-2016-makes,
  habernal-gurevych-2016-argument,
  quensel-etal-2025-investigating}. Other subtasks of the field
include detecting argument constituents
\citep{teufel-etal-2009-towards} and claims
\citep{wuhrl-klinger-2021-claim}; fact-checking of claims
\citep{thorne-vlachos-2018-automated}, reconstructing their structure
\citep{li-etal-2022-neural} and linking them
\citep{ebner-etal-2020-multi}. A prominent task is assessing the
quality of arguments \citep{wachsmuth-etal-2024-argument}, which
includes quantifying the logical, rhetorical, and dialectical quality
of the argument
\citep{Blair2011-BLAGIT}. \citet{quensel-etal-2025-investigating} use
regression analysis to investigate the subjective factors of emotions,
storytelling, and hedging and their impact on argument strength,
finding that the influence of the emotion depends on the rhetorical
utilization with respect to the argument quality.

Most relevant to our work is the subtask of assessing an argument's
convincingness. Different dimensions influence an argument's
convincingness, where textual qualities play a role
\citep{habernal-gurevych-2016-makes,habernal-gurevych-2016-argument}
as well as the personalities of the receivers
\citep{lukin-etal-2017-argument,
  al-khatib-2020-characteristics-debaters-persuasiveness}. Especially
important for argument convincingness are emotions -- their importance
has been demonstrated in the fields of computational argumentation
\citep{habernal-gurevych-2016-argument, habernal-gurevych-2016-makes,
  wachsmuth-etal-2017-computational, greschner-klinger-2025-fearful}, philosophy \citep{aristotle1991rhetoric}, and psychology \citep{bohner_affect-persuasion-1992, petty_mood_persuasion_1993,
  pfau_affect_resistance_2006, Worth1987CognitiveMO,Benlamine2015EmotionsArgumentesEmpirical}. In the field of natural
language processing (NLP), prior work treats emotions in arguments as
a binary variable, as one of many factors of convincingness
\citep{habernal-gurevych-2016-argument}, or rate the emotional appeal
\citep{wachsmuth-etal-2017-computational,lukin-etal-2017-argument} of
the argument. Emotions in arguments are frequently treated as a
fallacy \citep{jin-etal-2022-logical, ziegenbein-etal-2023-modeling},
recently, a study employs LLMs to inject emotional appeals into
fallacious arguments, finding that emotional framing reduces human
fallacy detection and that fear, sadness, and enjoyment significantly
increase perceived convincingness compared to neutral states
\citep{chen2025emotionallychargedlogicallyblurred}.

The most relevant works for our study include
\citet{chen-eger-2025-emotions}, who examine emotion intensity's
impact on convincingness through LLM-based manipulation, and
\citet{greschner-klinger-2025-fearful}, who focus on discrete emotion
categories in German arguments. However, both studies treat emotions
as categorical labels rather than exploring the underlying cognitive
processes that generate these emotional responses. Our work differs
fundamentally because we investigate whether appraisal theories --
which model the subjective cognitive evaluation leading to emotions --
provide sufficient explanations for convincingness compared to
categorical emotion models. 

\begin{table}
  \centering\small
  \begin{tabularx}{1.0\linewidth}{l>{\setlength{\baselineskip}{.5\baselineskip}}X}
    \toprule
    Conf. & Prompt Part \\
    \cmidrule(r){1-1}\cmidrule(l){2-2}
    \multirow{5}{*}{\rotatebox{90}{Prefix}}%
                  & \texttt{You are an expert on annotating argumentative texts. You will have to solve the following tasks. 
                    Task: Convincingness Prediction: You will be given
                    an argumentative text. Your task is to assign how
                    convincing a person
                    would find the argument} \\
    \cmidrule(r){1-1}\cmidrule(l){2-2}
    \multirow{3}{*}{\rotatebox{90}{Emo$\rightarrow$}} & \texttt{given that they felt the emotion \{emotion\} after hearing the argument} \\
    \cmidrule(r){1-1}\cmidrule(l){2-2}
    \multirow{3}{*}{\rotatebox{90}{Appr$\rightarrow$}} & \texttt{given that they assigned the following appraisals after hearing the argument:
\{appraisals\}} \\
    \cmidrule(r){1-1}\cmidrule(l){2-2}
    \multirow{5}{*}{\rotatebox{90}{CVC}}& 
\texttt{on a 1-5 scale.
           Rating scale:
           1 = Not at all convincing
           2 = Slightly convincing 
           3 = Moderately convincing
           4 = Very convincing
           5 = Extremely convincing
           Argument: "{argument}"
           Respond with valid JSON containing only the numerical rating:
           \{\{"rating": [number from 1-5]\}\}} \\
    \bottomrule
  \end{tabularx}
  \caption{Prompts for the Pipeline configuration of appraisal/emotion
    conditioned convincingness prediction. The Emo and Appr sections
    are optionally added to the convincingness prediction as parameters.}
  \label{tab:pipeline_prompts}
\end{table}

\begin{table*}
  \centering\small

  \begin{tabularx}{1.0\linewidth}{l>{\setlength{\baselineskip}{.5\baselineskip}}X}
    \toprule
    Conf. & Prompt Part \\
    \cmidrule(r){1-1}\cmidrule(l){2-2}
    \multirow{2}{*}{\rotatebox{90}{Prefix}}%
                  & \texttt{You are an expert on annotating argumentative texts. You will have to solve the following tasks.} \\
    \cmidrule(r){1-1}\cmidrule(l){2-2}
    \multirow{5}{*}{\rotatebox{90}{Emo$\leftrightarrow$CVC}} & \texttt{Task: Emotion Prediction. You will be given an argumentative text. Your task is 
to assign the strongest emotion that is evoked in a person hearing the argument. 
The emotion categories to choose from are: anger, disgust, fear, guilt, joy, pride, relief, sadness, shame, surprise, trust.
\newline
Task 2: Convincingness Prediction.
Your task is to assign how convincing a person would find the argument on a 1-5 scale.
Rating scale:
1 = Not at all convincing
2 = Slightly convincing  
3 = Moderately convincing
4 = Very convincing
5 = Extremely convincing
Argument: "{argument}".
Respond with valid JSON containing the emotion and the numerical rating:
\{\{"emotion": "emotion\_name", "rating": [number from 1-5]\}\}
} \\
    \cmidrule(r){1-1}\cmidrule(l){2-2}
    \multirow{6}{*}{\rotatebox{90}{Appr$\leftrightarrow$CVC}} & \texttt{Task: Appraisal Prediction: You will be given an argumentative text. 
Your task is to label each appraisal dimension on a 1-5 scale from the perspective of a person hearing the argument.
For each appraisal, one means the appraisal does not apply at all, 5 means it applies extremely.
The appraisals are:
Suddenness: the argument appears sudden or abrupt to the receiver
Suppression: the receiver tries to shut the argument out of their mind
Familiarity: the argument is familiar to the receiver
Pleasantness: the argument is pleasant for the receiver
Unpleasantness: the argument is unpleasant for the receiver
Consequencial Importance: the argument has important consequences for the receiver
Positive Consequentiality: the argument has positive consequences for the receiver
Negative Consequentiality: the argument has negative consequences for the receiver
Consequence Manageability: the receiver can easily live with the unavoidable consequences of the argument
Internal Check: the consequences of the argument clash with the receiver’s standards and ideals
External Check: the consequences of the argument violate laws or socially accepted norms
Response urgency: the receiver urges to immediately respond to the argument
Cognitive Effort: processing the argument requires a great deal of energy of the receiver
Argument Internal Check: statements in the argument clash with the receiver’s standards and ideals
Argument External Check: statements in the argument violate laws or socially accepted norms
\newline
Task 2: Convincingness Prediction.
Your task is to assign how convincing a person would find the argument on a 1-5 scale.
Rating scale:
1 = Not at all convincing
2 = Slightly convincing  
3 = Moderately convincing
4 = Very convincing
5 = Extremely convincing
Argument: "\{argument\}".
You must respond with ONLY a valid JSON object. Each key must have an integer value between 1 and 5. Format:
\{\{
  "suddenness": 1,
  "suppression": 1,
  "familiarity": 1,
  "pleasantness": 1,
  "unpleasantness": 1,
  "consequential\_importance": 1,
  "positive\_consequentiality": 1,
  "negative\_consequentiality": 1,
  "consequence\_manageability": 1,
  "internal\_check": 1,
  "external\_check": 1,
  "response\_urgency": 1,
  "cognitive\_effort": 1,
  "argument\_internal\_check": 1,
  "argument\_external\_check": 1,
  "convincingness": 1
\}\}.
Replace each "1" with your actual rating (1-5) for that dimension.} \\
    \bottomrule
  \end{tabularx}
  \caption{Prompts for the Joint configuration of appraisal/emotion
    conditioned convincingness prediction.}
  \label{tab:joint_prompts}
\end{table*}

\section{Methods}
The goal of our study is to understand if appraisals and/or emotion
categories help to computationally assess an arguments
convincingness. We do so with prompting experiments across a set of
language models. It is important to keep in mind that we consider all
three tasks (appraisal prediction, emotion prediction, and
convincingness prediction) to be subjective.

\subsection{Prompting Setup}
We compare a set of prompting configurations, which we explain in the
following. Since prior work highlighted the complexity of prompt
formulations and considering that
\citet{greschner-klinger-2025-fearful} demonstrate minimal impact of
prompting methods (zero-shot, few-shot, chain-of-thought) on emotion
predictions, we opt to focus exclusively on zero-shot prompts in our
study. This approach ensures a straightfoward, plug-and-play
implementation while avoiding confounding effects of complex prompting
strategies.

\begin{table*}[]
    \centering\small\setlength{\tabcolsep}{3pt} 
\begin{tabularx}{\textwidth}{X cl rrrrrrrrrrrrrrr}
\toprule
    Argument & \rotatebox{0}{CVC} & Emotion & \rotatebox{90}{Suddenness}& \rotatebox{90}{Suppression}& \rotatebox{90}{Familiarity}& \rotatebox{90}{Pleasantness}& \rotatebox{90}{Unpleasantness}& \rotatebox{90}{Consequencial Importance}& \rotatebox{90}{Positive Consequentiality}& \rotatebox{90}{Negative Consequentiality}& \rotatebox{90}{Consequence Manageability}& \rotatebox{90}{Internal Check}& \rotatebox{90}{External Check}& \rotatebox{90}{Response Urgency}& \rotatebox{90}{Cognitive Effort}& \rotatebox{90}{Argument Internal Check}& \rotatebox{90}{Argument External Check} \\
    \cmidrule(r){1-1}\cmidrule(lr){2-2}\cmidrule(lr){3-3}\cmidrule(r){4-4}\cmidrule(r){5-5}\cmidrule(r){6-6}\cmidrule(r){7-7}\cmidrule(r){8-8}\cmidrule(r){9-9}\cmidrule(r){10-10}\cmidrule(r){11-11}\cmidrule(r){12-12}\cmidrule(r){13-13}\cmidrule(r){14-14}\cmidrule(r){15-15}\cmidrule(r){16-16}\cmidrule(r){17-17}\cmidrule(l){18-18}
    \multirow{5}{=}{it could be considered that holocaust denial is a hate crime, laws should be in place to protect the memory of those who have perished in unspeakable crimes such as the holocaust.} & 5 & Relief & 1 & 1 & 3 & 3 & 1 & 2 & 4 & 1 & 4 & 1 & 1 & 1 & 5 & 1 & 1 \\
     & 4 & Trust & 3 & 1 & 4 & 1 & 4 & 1 & 1 & 1 & 3 & 3 & 2 & 2 & 5 & 1 & 4 \\
     & 3 & Sadness & 1 & 2 & 3 & 1 & 4 & 2 & 1 & 1 & 1 & 2 & 2 & 2 & 2 & 1 & 3 \\
     & 4 & Anger & 1 & 1 & 3 & 1 & 3 & 1 & 1 & 1 & 4 & 1 & 1 & 1 & 5 & 2 & 1 \\
     & 4 & Trust & 1 & 1 & 3 & 3 & 1 & 4 & 3 & 1 & 4 & 1 & 1 & 4 & 5 & 2 & 1 \\
    \cmidrule(r){1-1}\cmidrule(lr){2-2}\cmidrule(lr){3-3}\cmidrule(r){4-4}\cmidrule(r){5-5}\cmidrule(r){6-6}\cmidrule(r){7-7}\cmidrule(r){8-8}\cmidrule(r){9-9}\cmidrule(r){10-10}\cmidrule(r){11-11}\cmidrule(r){12-12}\cmidrule(r){13-13}\cmidrule(r){14-14}\cmidrule(r){15-15}\cmidrule(r){16-16}\cmidrule(r){17-17}\cmidrule(l){18-18}
    \multirow{5}{=}{it is a great thing when dads want to stay home with their children, but often they are needed as the main income earner so subsidizing them would help greatly.} & 1 & Sadness & 3 & 5 & 4 & 1 & 5 & 2 & 5 & 1 & 5 & 1 & 5 & 5 & 5 & 5 & 5 \\
     & 3 & Relief & 1 & 1 & 2 & 1 & 2 & 2 & 1 & 1 & 4 & 2 & 2 & 2 & 5 & 1 & 2 \\
     & 4 & Sadness & 1 & 1 & 4 & 4 & 1 & 2 & 1 & 3 & 1 & 4 & 1 & 2 & 2 & 1 & 1 \\
     & 1 & Disgust & 3 & 5 & 1 & 1 & 5 & 2 & 2 & 1 & 4 & 1 & 5 & 1 & 2 & 4 & 5 \\
     & 4 & Joy & 1 & 1 & 3 & 3 & 1 & 2 & 1 & 1 & 1 & 4 & 1 & 2 & 2 & 1 & 1 \\
    \bottomrule
\end{tabularx}
    \caption{Examples from the \contarga corpus. Emotion, appraisal, and convincingness assessments are from five individual
      annotators. The convincingness (CVC) and each appraisal dimension are evaluated on a 1--5 scale.}
    \label{tab:contarga_examples}
\end{table*}

\paragraph{Single Model.}
To create a baseline, we prompt the model to predict the convincingness of a given argument (we refer to this task as \textit{CVC} prediction). Similarly, we use a plain zero-shot prompting setting to predict (1) the emotion categories and (2) the appraisal dimensions as a single task.

\paragraph{Pipeline Model.} 
We expand two configurations, one in
which we add the appraisal information (\textit{Appr$\rightarrow$CVC}, and one in which we add emotion information to the prompt (\textit{Emo$\rightarrow$CVC}). 
In these two
cases, the appraisal and emotion information stems from the annotated
data, and may differ for otherwise textually comparable instances (see
Section~\ref{sec:data} for an explanation of the data we use). These two settings are used to compare the performance to the baseline CVC prediction. This
setup only allows a unidirectional information flow from the
emotion/appraisal representation to the convincingness
prediction. Table~\ref{tab:pipeline_prompts} shows the prompts for the
pipeline configuration.

\paragraph{Joint Model.} Presumably, the convincingness of an argument
does not only depend on the appraisal, but also the other way
around. Therefore, we also perform a joint model experiment, in which
the language model is requested to output emotion or appraisal
variables, together with the convincingness. We refer to this model as
\textit{Appr$\leftrightarrow$CVC} and
\textit{Emo$\leftrightarrow$CVC}. This setup enables an information
flow between the emotion representation and the convincingness
assessment in both directions.

\subsection{Models}
We prompt three large language models (LLMs), namely
Mistral-Small-2407 (\mistral), LLaMA3.3:70B (\llama) and
Gemma-3-27B-IT (\gemma). Mistral-Small-2407 \citep{jiang2023mistral7b} is a compact decoder-only
transformer with 22 billion parameters, optimized for fast inference
and low latency applications. LLaMA3.3:70B \citep{grattafiori2024llama3herdmodels} is a large-scale generative
model with 70 billion parameters, designed for high-performance
instruction following and multilingual reasoning. Gemma-3-27B-IT \citep{gemmateam2025gemma3technicalreport} is an
instruction-tuned model with 27 billion parameters, optimized for task
completion. The Mistral and Gemma models are accessed via their
respective APIs. The LLaMA model is accessed locally via Ollama\footnote{\url{https://ollama.com/}}. For
all models, we set the temperature to 0.1 and leave all other parameters at their respective default values\footnote{Code for all experiments: \url{https://github.com/LynnGreschner/categorical-emotions-or-appraisals}}.

\section{Data}
\label{sec:data}
The data we use in our study is the \contarga data set. It has been
presented as part of the Contextualized Argument Appraisal Framework,
in which the authors propose a set of concrete appraisal variables
that may be used for the evaluation of arguments
\citep{greschner2025trustmeiconvince}.  The corpus situates argument
appraisal in its communicative context, capturing the interplay
between sender, receiver, and argument rather than treating persuasion
as a purely textual property. \contarga comprises 800 arguments
drawn from the UKPConvArgv1 \citep{habernal-gurevych-2016-argument} and
IBM-Rank-30k \citep{gretz-etal-2019-IBMcorpus} datasets, each
annotated by five participants, resulting in 4,000 contextualized
annotations. Table~\ref{tab:contarga_examples} displays examples of two arguments with convincingness, emotion, and appraisal assessments from 5 individual annotations each.

An important challenge to obtain subjective labels such as emotions,
appraisals or convincingness is to access a person's evaluation in a
realistic setup, without using external annotators that would need to
reconstruct a presumable evaluation of somebody else.  Data were
therefore collected in a role-playing setup which simulated a
town-hall meeting in which participants engaged with arguments on 39
topics, then reported their emotional response, cognitive appraisal,
and perceived convincingness. Annotations include discrete emotion
categories (e.g., anger, trust, relief, sadness), intensity ratings,
free-text emotion causes, and 15 appraisal dimensions (such as
familiarity, pleasantness, response urgency, and perceived
consequences). Participants also provided demographic information and
Big Five personality traits for both themselves (as receivers) and
their imagined argument sender.

The statistical analysis of the data revealed correlations between emotional
and cognitive dimensions: positive emotions (trust, pride, joy,
relief) correlate with high convincingness, while negative
emotions (anger, disgust, sadness) correspond to low
convincingness. Appraisal variables such as pleasantness, positive
consequentiality, and familiarity further enhance convincingness, whereas
unpleasantness and norm violation decrease it.

The authors of the paper do, however, not
conduct a modeling study which would allow any insights in the role of
convincingness modeling under a particular argument appraisal or
emotion. Therefore, \contarga constitutes a good starting point for
our study.

\section{Experiments}
We now turn to the experimental settings of our experiments and answer each research question.

\subsection{RQ1: Which emotion model helps LLMs to improve convincingness predictions?}

We aim at understanding if providing information about the evoked emotion of a given argument improves the performance of LLMs on the convincingness prediction task. More specifically, we investigate whether there is a difference in the emotion model that provides the information about the emotion. To this end, we compare providing discrete emotion categories (e.g., anger, joy, fear, \ldots) and appraisal dimensions (familiarity, suddenness, cognitive effort, \ldots) to the model for the convincingness prediction task.

\paragraph{Experimental Setting.}
We prompt the three LLMs to predict the convincingness of a given argument on a 1--5 scale, which serves as the baseline for the task. It is a zero-shot prompt, the exact phrasing is displayed in Table~\ref{tab:pipeline_prompts}. For investigating the effect of providing the discrete emotion category, the LLM is provided with the dominant emotion that was evoked in a receiver of a given argument. Similarly, in the third task, we provide the appraisal values for all 15 appraisal dimensions that a participant annotated. The exact prompts used for the experiments can again be found in Table~\ref{tab:pipeline_prompts}. Each model is prompted up to four times if no valid prediction can be extracted\footnote{For 75 instances ($\sim$2\% of the data), the model fails to provide a valid answer even after the fourth attempt. Such cases are excluded from the evaluation of all tasks.}. Due to the instability of LLM responses, we run all experiments five times and report the average performance across runs to ensure robust and reliable results.

\paragraph{Results.}\label{sec:rq1_results}
Table~\ref{tab:mainresults} displays the results of the convincingness prediction task using the different settings. \mistral and \gemma perform best on the convincingness prediction task (.33), \llama performs worst with .27. Compared to this baseline, providing the discrete emotion category that was evoked in a receiver of the argument improves the convincingness prediction for all models. The strongest improvement (+.09) is observed for \llama, whereas \gemma then performs best across models with .41.
Interestingly, providing the appraisal values does not improve the convincingness predictions of \gemma; the model even performs worse than the baseline. However, both \mistral and \llama perform best when being provided with the appraisal dimensions (.41 and .42, respectively).

With respect to our research question, we observe that information about the emotion improves the CVC prediction of all models. While the discrete emotion category reliably improves the performance, we see stronger improvements from the appraisal dimensions, even though it fails for one model. Our results suggest that, while emotional context generally improves the prediction of persuasiveness, different models vary in their sensitivity to categorical and dimensional approaches to representing emotions.

\begin{table}
  \centering
  \begin{tabular}{l lll}
    \toprule
    & \multicolumn{3}{c}{CVC Spearman's $\rho$} \\
    \cmidrule(l){2-4}
    Config. & Mistral & Gemma & Llama \\
    \cmidrule(r){1-1}\cmidrule(lr){2-2}\cmidrule(lr){3-3}\cmidrule(l){4-4}
    CVC Basel. & .33 & .33 & .27  \\
    Emo$\rightarrow$CVC & .38\gd{.05} & .41\gd{.08} & .36\gd{.09} \\
    Appr$\rightarrow$CVC & .41\gd{.08} & .24\rd{.09} & .42\gd{.15} \\
    \cmidrule(r){1-1}\cmidrule(lr){2-2}\cmidrule(lr){3-3}\cmidrule(l){4-4}
    Emo$\leftrightarrow$CVC & .31\rd{.02} & .25\rd{.08} & .29\gd{.02} \\
    Appr$\leftrightarrow$CVC & .32\rd{.01} & .30\rd{.03} & .32\gd{.05} \\
    \bottomrule
  \end{tabular}
  \caption{Main results of the convincingness (CVC) prediction task, comparing the Pipeline and Joint Setup across three language models. The CVC correlations are
    reported in Spearman's $\rho$. Differences of each score compared to the CVC baseline are shown with $\Delta$ values.}
  \label{tab:mainresults}
\end{table}

\begin{table*}
\centering\small\setlength{\tabcolsep}{5pt}
\begin{tabular}{l ccc ccc ccc ccc ccc ccc}
\toprule
& \multicolumn{9}{c}{Single} & \multicolumn{9}{c}{Joint}\\
\cmidrule(r){2-10}\cmidrule(l){11-19}
& \multicolumn{3}{c}{Mistral} & \multicolumn{3}{c}{Llama} & \multicolumn{3}{c}{Gemma} & \multicolumn{3}{c}{Mistral} & \multicolumn{3}{c}{Llama} & \multicolumn{3}{c}{Gemma} \\
\cmidrule(r){2-4}\cmidrule(lr){5-7}\cmidrule(lr){8-10}\cmidrule(lr){11-13}\cmidrule(lr){14-16}\cmidrule(l){17-19}
 & P & R & \F & P & R & \F & P & R & \F & P & R & \F & P & R & \F & P & R & \F \\
\cmidrule(r){2-2}\cmidrule(lr){3-3}\cmidrule(lr){4-4}\cmidrule(lr){5-5}\cmidrule(lr){6-6}\cmidrule(lr){7-7}\cmidrule(lr){8-8}\cmidrule(lr){9-9}\cmidrule(lr){10-10}\cmidrule(lr){11-11}\cmidrule(lr){12-12}\cmidrule(lr){13-13}\cmidrule(lr){14-14}\cmidrule(lr){15-15}\cmidrule(lr){16-16}\cmidrule(lr){17-17}\cmidrule(lr){18-18}\cmidrule(l){19-19}
Anger & .18 & .52 & .26 & .19 & .43 & .26  & .18 & .25 & .20 & .18 & .44 & .26 & .20 & .41 & .27  & .19 & .23 & .21   \\
Disgust & .10 & .09 & .09 & .13 & .12 & .12  & .09 & .08 & .08 & .10 & .12 & .11 & .14 & .09 & .11  & .09 & .09 & .09  \\
Fear & .06 & .32 & .11 & .08 & .26 & .12  & .05 & .36 & .09 & .06 & .34 & .11 & .08 & .27 & .12  & .05 & .40 & .09  \\
Guilt & .00 & .00 & .00 & .07 & .14 & .09  & .09 & .10 & .10 & .17 & .01 & .02 & .05 & .16 & .08  & .08 & .12 & .09  \\
Joy & .14 & .31 & .19 & .20 & .11 & .14  & .20 & .08 & .12 & .15 & .28 & .19 & .18 & .08 & .11  & .19 & .09 & .13  \\
Pride & .10 & .20 & .13 & .10 & .12 & .11  & .15 & .04 & .06 & .12 & .18 & .14 & .13 & .12 & .12  & .15 & .05 & .08  \\
Relief & .22 & .02 & .03 & .17 & .05 & .07  & .16 & .06 & .08 & .12 & .00 & .00 & .19 & .05 & .08  & .16 & .03 & .06  \\
Sadness & .37 & .13 & .20 & .32 & .23 & .27  & .27 & .23 & .25 & .36 & .15 & .21 & .31 & .21 & .25  & .29 & .22 & .25  \\
Shame & .19 & .01 & .02 & .05 & .00 & .00  & .14 & .03 & .05 & .20 & .03 & .02 & .00 & .00 & .00  & .18 & .07 & .10  \\
Surprise & .36 & .02 & .04 & .29 & .03 & .05  & .31 & .04 & .08 & .37 & .03 & .05 & .30 & .04 & .07  & .31 & .04 & .08  \\
Trust & .28 & .14 & .19 & .27 & .33 & .30  & .24 & .39 & .30 & .27 & .25 & .26 & .27 & .39 & .32  & .25 & .42 & .31  \\
\cmidrule(r){2-2}\cmidrule(lr){3-3}\cmidrule(lr){4-4}\cmidrule(lr){5-5}\cmidrule(lr){6-6}\cmidrule(lr){7-7}\cmidrule(lr){8-8}\cmidrule(lr){9-9}\cmidrule(lr){10-10}\cmidrule(lr){11-11}\cmidrule(lr){12-12}\cmidrule(lr){13-13}\cmidrule(lr){14-14}\cmidrule(lr){15-15}\cmidrule(lr){16-16}\cmidrule(lr){17-17}\cmidrule(lr){18-18}\cmidrule(l){19-19}
Avg & .18 & .16 & .12  & .17 & .16 & .14  & .17 & .15 & .13 & .19 & .16 & .12 & .17 & .17 & .14  & .18 & .16 & .13  \\
\bottomrule
\end{tabular}
\caption{Precision, Recall, and F1 values (macro-average) for the emotion classification task, comparing single emotion predictions and the emotion predictions from the joint modeling task.}
\label{tab:emotion_classwise}
\end{table*}

\subsection{RQ2: Does jointly predicting emotions/appraisals and convincingness improve the performance compared to the single task predictions?}

While \citet{greschner2025trustmeiconvince} report strong correlations between emotions, appraisals and convincingness, the convincingness of an argument does presumably not merely depend on the emotion and appraisal, but these evoked emotions also depend on the convincingness. The following experiment examines whether the joint prediction of emotions, appraisals, and convincingness improves the convincingness predictions.

\paragraph{Experimental Setting.}
We prompt the three LLMs to jointly predict emotions/appraisals and
convincingness. In contrast to our first experiment, here the models
do not get the information about the emotions/appraisals, but have to
jointly predict them simultaneously with the convincingness. We chose this setup due to presumably bi-directional effects between emotions/appraisals and convincingness. In addition, predicting the variables jointly is a more realistic setup since most corpora do not have emotion/appraisal labels in addition to convincingness ratings.
The exact prompt formulations can be found in Table~\ref{tab:joint_prompts}.

\paragraph{Results.}
We report the results of this experiment in Table~\ref{tab:mainresults}. Jointly predicting emotions and convincingness only improves the convincingness prediction for \llama (+.05). Both \mistral and \gemma perform worse than the baseline in the joint prediction setting. Similarly, the joint prediction of appraisals and convincingness does not improve the prediction of the convincingness compared to the baseline, except for \llama, which also shows the best overall performance in the joint setting (.32). 

However, while \mistral and \gemma fail to beat the baseline in the joint setting, we do observe better performance of all models for jointly predicting appraisals and convincingness compared to jointly predicting discrete emotions and convincingness.

\section{Analysis of Emotion Models}

We now turn to the two different emotion models in more detail. We aim
at understanding why and how the models improve the convincingness prediction.

\paragraph{Emotions.}

We display the performance of all three models on the single and joint emotion classification task in Table~\ref{tab:emotion_classwise}. Overall, the models show low performance on the emotion classification task in both the single (averaged \F scores of .12, .14, .13 for \mistral, \llama, \gemma, respectively) and the joint setting (averaged \F scores of .12, .14, .13 for \mistral, \llama, \gemma, respectively). There are minor differences in the precision and recall values when comparing the settings.

Taking a closer look at the class-wise performance, we find that negative emotions (\anger, \fear) show high recall but low precision values, in line with results of emotion predictions on German arguments \citep{greschner-klinger-2025-fearful}. 
Overall, all models perform best on predicting negative emotions (\anger, \disgust, \fear, \sadness), with the exception of one positive emotion (\trust). Trust is the most frequent emotion label in the gold data.

Turning to model-specific performance, we find that \llama consistently achieves the best performance in both single and joint settings, particularly high results are seen for predicting \trust (.30 and .32 \F in the single and joint settings, respectively) and \sadness (.25 and .27 \F). In contrast, \mistral benefits from the joint modeling, showing improved performance across several emotion categories, notably for \anger and \sadness. For \gemma, we see unique behavior, on the one hand demonstrating strong capabilities for predicting \trust (.30 and .31 F), but struggling with \joy and \pride (compared to the other models). Notably, \gemma is the only model that does not benefit from joint modeling, and as described in Section~\ref{sec:rq1_results}, the CVC prediction of \gemma fails to improve even when provided with gold-label emotion information, suggesting fundamental differences in how this model processes and integrates emotion information.

\paragraph{Appraisals.}
The performance of all models on the appraisal prediction task is low -- in both the single and joint setting. All models predict the appraisal values similarly when comparing the single and joint predictions. However, there are some model-specific and appraisal-specific differences.

We find the best performance for predicting appraisals for the appraisal dimensions of \pleasantness (correlation values of .32, .31, .31 for \mistral, \llama, \gemma, respectively on the single task and .32, .32, .32 on the joint task) and \unpleasantness (.28, .29, .29 for \mistral, \llama, \gemma, respectively on the single task and .28, .30, .30 on the joint task). The appraisal dimension of \familiarity also shows comparably high results (.20, .12, .19 for \mistral, \llama, \gemma, respectively on the single task and .20, .11, .19 on the joint task). The appraisal dimensions \negconseq and \conseqmanage show the lowest performance.

Considering model-specific behaviour, we see that for the two best-performing appraisals (\pleasantness and \unpleasantness), all three models perform on par or marginally better in the joint setting. However, for \familiarity, only \llama struggles with the prediction in both settings (.12 and .11 in single and joint settings, respectively).

In contrast to the CVC prediction, where the single task predictions (i.e., the CVC baseline) show higher performance compared to the joint prediction of CVC and appraisals, there is only a marginal difference between using a single or joint setting when predicting appraisal dimensions in a zero-shot prompting setting.

\subsection{Discussion}

The most significant finding is that appraisal dimensions consistently outperform categorical emotions in improving convincingness predictions. This aligns with appraisal theory's premise that cognitive evaluations are more predictive than discrete emotional categories. The superior performance is particularly evident in the pipeline setting, where gold-standard appraisal information yield the strongest improvements (up to +.15 for \llama). However, while the prediction of convincingness performs moderately, the performance of all models on the appraisal prediction task is low. Our findings of consistent underperformance indicate that zero-shot prompting methods may be insufficient for capturing the complex bidirectional relationships between emotions, appraisals, and convicingness. This highlights the need for more distinguished appraisal predictors for automatically predicting convincingness in arguments.

\section{Conclusion}

In this work, we investigated whether appraisal theories can computationally explain argument convincingness compared to categorical emotion models. Using zero-shot prompting experiments of \gemma, \llama, and \mistral on the \contarga corpus, our results demonstrate that information from both emotion models improves the convincingness prediction task over the baseline model. Categorical emotions improve performance consistently across models; however, appraisal dimensions showed stronger effects on the predictions. This supports our hypothesis that the subjective nature of argument evaluation benefits from the more granular cognitive assessment provided by appraisal dimensions. Therefore, moving beyond categorical emotion models toward cognitively grounded appraisal frameworks can enhance our understanding of subjective argument evaluation.  

\begin{table}
\centering\small\setlength{\tabcolsep}{2.5pt} 
\begin{tabular}{l rrr rrr}
\toprule
& \multicolumn{3}{c}{Single} & \multicolumn{3}{c}{Joint}\\
\cmidrule(lr){2-4}\cmidrule(l){5-7}
& M & L & G & M & L & G \\
\cmidrule(r){1-1}\cmidrule(l){2-2}\cmidrule(l){3-3}\cmidrule(l){4-4}\cmidrule(l){5-5}\cmidrule(l){6-6}\cmidrule(l){7-7}
Arg. Ext. Check  & .09 & .13 & .06 & .10 & .13 & .07  \\
Arg. Int. Check  & .11 & .11 & .10 & .11 & .10 & .10  \\
Cog. Eff.  & .03 & .01 & .02 & .01 & .01 & .00  \\
Conseq. Manag.  & $-$.02 & $-$.02 & $-$.02 & $-$.01 & $-$.01 & $-$.01 \\
Conseq. Import.  & .05 & .04 & .03 & .05 & .04 & .03  \\
Ext. Check  & .15 & .14 & .13 & .15 & .14 & .12  \\
Fam.  & .20 & .12 & .19 & .21 & .11 & .19  \\
Int. Check  & .03 & .02 & .01 & .03 & .03 & .00 \\
Neg. Consequ.  & $-$.03 & $-$.05 & $-$.02 & $-$.03 & $-$.04 & $-$.02  \\
Pleas  & .32 & .31 & .31 & .32 & .32 & .32  \\
Pos. Consequ.  & .12 & .13 & .12 & .12 & .13 & .12  \\
Resp. Urg.  & .10 & .14 & .13 & .12 & .13 & .11 \\
Sudd.  & .08 & .15 & .07 & .15 & .16 & .07  \\
Supp.  & .13 & .12 & .08 & .14 & .11 & .09  \\
Unpleas.  & .28 & .29 & .29 & .28 & .30 & .30 \\

\cmidrule(r){1-1}\cmidrule(l){2-2}\cmidrule(l){3-3}\cmidrule(l){4-4}\cmidrule(l){5-5}\cmidrule(l){6-6}\cmidrule(l){7-7}
Average  & .11 & .11 & .09 & .11 & .11 & .10\\
\bottomrule
\end{tabular}
\caption{Appraisal Prediction Evaluation Results. Results of the appraisal prediction task for individual appraisal dimensions and the average, all reported as correlations using Spearman's $\rho$.}
\label{tab:appraisal_evaluation}
\end{table}

Our study reveals clear challenges. The joint modeling of
emotions/appraisals and convincingness, while theoretically promising
(allowing bidirectional information flow), did not yield improvements
over pipeline approaches using gold-standard appraisal information to
guide convincingness predictions. This underlines the need for
developing better predictors for both emotion and appraisal dimensions
in the context of arguments -- potentially through fine-tuning, richer
context modeling, or multi-task learning strategies.

Future work should focus on advancing automatic appraisal and emotion prediction models in argumentation settings, exploring adaptive and fine-tuned approaches, and validating appraisal-based models across languages and domains. Moreover, we lay the ground work for integrating appraisal information into downstream applications such as argument generation. Adopting cognitively grounded emotion models offers the potential to make computational argument analysis both more robust and more human-aligned.

\section{Limitations}
The following limitations should be considered when interpreting our results. With respect to our methodology, we do not explore few-shot learning, chain-of-thought reasoning, or fine-tuning approaches. We focus on zero-shot prompting, which possibly does not capture the full spectrum of LLM capabilities for convincingness prediction, because (1) and (2) adopting this straightforward approach allows us to clearly assess the validity of appraisals in arguments rather than confounding the results with the complexity of more elaborate prompting or fine-tuning strategies. The generally low performance on emotion and appraisal prediction, however, rather reflects the difficulty of such subjective tasks rather than fundamental flaws in our approach. Our evaluation focuses primarily on correlation metrics, which do effectively capture relationships between emotions and convincingness, but might not fully reveal nuanced patterns in model behavior. We leave extensive error analysis to understand the underlying reasons for the model's performance variability to future work.

All experiments are conducted using the \contarga corpus, which, despite its careful construction, is limited to short, isolated arguments. The results of our experiments might differ when investigating arguments in context, i.e., in debates or social media discussions. Further, we only use English arguments, potentially missing cross-linguistic and cross-cultural variations in emotional responses to arguments.

\section{Ethical Considerations}

Our work has been approved by the ethics board of the University of Suttgart. Considering the ethical considerations our work poses, we follow the recommendations with respect to ethical challenges in emotion analysis by \citet{mohammad-2022-ethics-sheet}. We do not create any new artifacts other than automatic predictions. We use an existing, freely available dataset as our data source and use open-source large language models in our experiments. Automatically inferring emotional states using cognitive appraisals from argumentative texts could, in theory, provide insights into individuals' psychological states and cognitive processes that they might not have wanted to be inferred or analyzed. However, all participants involved in the creation of the data we use were informed and gave their consent for their answers to be used in scientific publications.

Automatic emotion analysis systems can be biased for various reasons \citep{kiritchenko-mohammad-2018-examining}. Our models may show biases related to demographics, cultural backgrounds, or linguistic expressions. Convincingness assessments are inherently subjective in nature, i.e., what makes an argument convincing varies for different people and groups. Predicting argument convincingness based on individual emotional and cognitive assessments could be misused to craft manipulative arguments that exploit emotional vulnerabilities or cognitive biases. In political, commercial, or propaganda contexts, such capabilities could undermine informed democratic deliberation. We recommend future work to include robust safety measurements against misuse for downstream tasks.

\section{Acknowledgements}
This project has been conducted as part of the \textsc{Emcona} (The Interplay of Emotions and Convincingness in Arguments) project, which is funded by the German Research Foundation (DFG, project KL2869/12--1, project number 516512112).

\section{Bibliographical References}
\label{sec:reference}

\bibliographystyle{lrec2026-natbib}
\bibliography{lit}

\bibliographystylelanguageresource{lrec2026-natbib}
\bibliographylanguageresource{languageresource}

\end{document}